\documentclass[10pt,twocolumn,letterpaper]{article}

\usepackage[pagenumbers]{cvpr} 

\definecolor{cvprblue}{rgb}{0.21,0.49,0.74}

\usepackage[table]{xcolor}
\definecolor{E6F5F0}{HTML}{E6F5F0}
\definecolor{darkF7E0D5}{RGB}{209,154,128}
\definecolor{F5F9FF}{HTML}{edf6ff}
\colorlet{Light}{E6F5F0} 
\newcommand{\CC}[1]{\cellcolor{Light}}

\newcommand{\demph}[1]{\textcolor{gray!80}{#1}}

\usepackage{tablefootnote}
\usepackage{booktabs}
\usepackage{fontawesome5}
\usepackage{graphicx}
\usepackage{multirow}
\usepackage{xcolor}
\usepackage{pifont}
\usepackage{amssymb}
\usepackage[pagebackref,breaklinks,colorlinks,allcolors=cvprblue]{hyperref}

\usepackage{pgfplots}
\usepackage{sansmath}
\pgfplotsset{compat=1.18}

\newcommand{\cmark}{\textcolor{green!60!black}{\ding{51}}}
\newcommand{\xmark}{\textcolor{red!70!black}{\ding{55}}} 
\newcommand\blfootnote[1]{
    \begingroup
    \renewcommand\thefootnote{}\footnote{#1} 
    \addtocounter{footnote}{-1}
    \endgroup
}

\def\paperID{-} 
\def\confName{CVPR}
\def\confYear{2026}

\def\ourmethod{Object-WIPER}
\def\ourbenchmark{WIPER-Bench}
\def\ourmetric{TokSim}

\newcommand{\nameemoji}{Object-WIPER\includegraphics[height=\baselineskip]{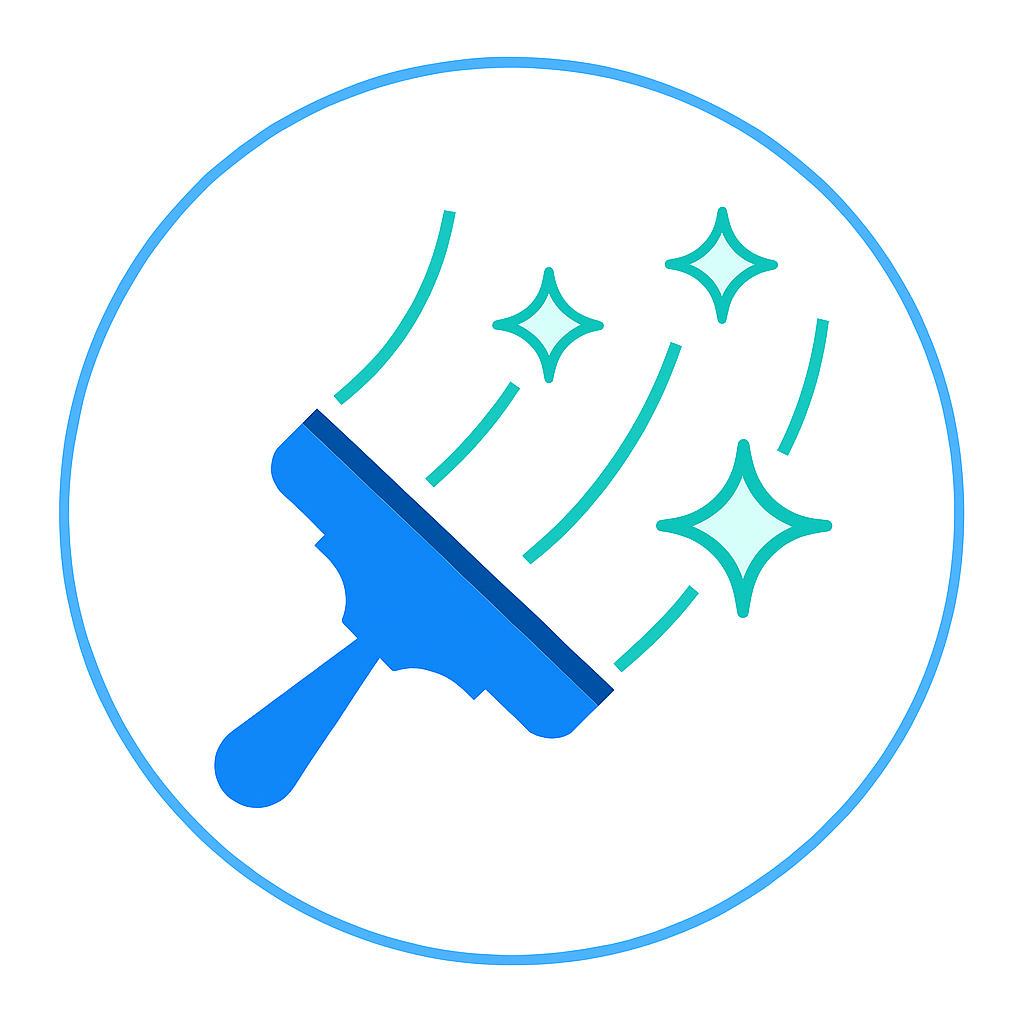}}

\usepackage{tocloft}

\title{\ourmethod: Training-Free Object and Associated Effect Removal in Videos}

\author{
Saksham Singh Kushwaha$^{1,*}$,
Sayan Nag$^{2}$, Yapeng Tian$^{1}$, Kuldeep Kulkarni$^{2}$\\
$^{1}$The University of Texas at Dallas, 
$^{2}$Adobe Research\\
\faGlobe\
\href{https://sakshamsingh1.github.io/object_wiper_webpage/}
{sakshamsingh1.github.io/object\_wiper\_webpage}
}

\begin{document}
\maketitle

\blfootnote{* Work done during an internship at Adobe.}

\addtocontents{toc}{\protect\setcounter{tocdepth}{-1}}

\begin{abstract}
In this paper, we introduce \textbf{\ourmethod}, a training-free framework for removing dynamic objects and their associated visual effects from videos, and inpainting them with semantically consistent and temporally coherent content. Our approach leverages a pre-trained text-to-video diffusion transformer (DiT). 
Given an input video, a user-provided object mask, and query tokens describing the target object and its effects, we localize relevant visual tokens via visual-text cross-attention and visual self-attention. This produces an intermediate effect mask that we fuse with the user mask to obtain a final foreground token mask to replace. We first invert the video through the DiT to obtain structured noise, then reinitialize the masked tokens with Gaussian noise while preserving background tokens. During denoising, we copy values for the background tokens saved during inversion to maintain scene fidelity. To address the lack of suitable evaluation, we introduce a new object removal metric that rewards temporal consistency among foreground tokens across consecutive frames, coherence between foreground and background tokens within each frame, and dissimilarity between the input and output foreground tokens. Experiments on \textsc{DAVIS} and a newly curated real-world associated effect benchmark (\textbf{\ourbenchmark}) show that \ourmethod~surpasses both training-based and training-free baselines in terms of the metric, achieving clean removal and temporally stable reconstruction without any retraining. Our new benchmark, source code, and pre-trained models will be publicly available.

\begin{figure}[t]
    \centerline{\includegraphics[width=\linewidth]{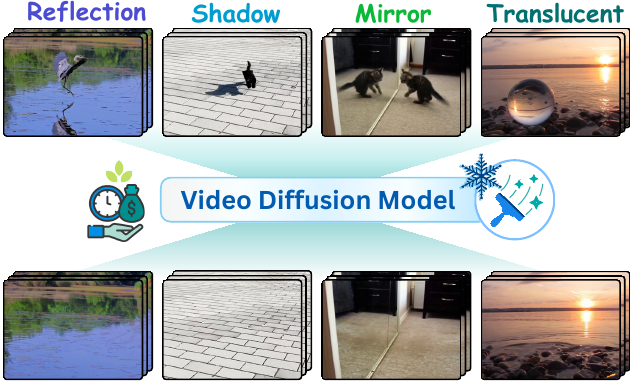}}
    \caption{\nameemoji{} removes undesired objects and their associated effects without any training, thereby avoiding substantial training time and computational resources.}
    \label{fig:teaser}
    \vspace{-5mm}
\end{figure}
\end{abstract}    
\section{Introduction}
\label{sec:intro}
Object removal from videos is an extremely important problem that has widespread applications like film and video production that require boom mic or crew removal, surveillance and privacy protection and creative content generation. The history of this problem has its genesis in the classical non-parametric video inpainting techniques \cite{barnes2009patchmatch, newson2014video, huang2016temporally, granados2012background} that use combinations of energy minimization, graph cuts and flow estimation techniques.  The video inpainting approaches evolved with the explosion of convolutional neural networks (CNNs) and recurrent neural networks (RNNs) in the past decade yielding superior results \cite{chang2019free, hu2020proposal, kim2019deep, xu2019deep, zhang2019internal}. The video inpainting approaches inherently focused only on filling the regions belonging to the object that is being removed through flow estimation techniques while completely ignoring the associated effects of the object like shadows and reflections. The very nature of these approaches lead to undue artifacts owing to the retention of the associated effects in the output videos. This drawback of retention of the associated effects is shared by the recent video inpainting methods \cite{bian2025videopainter, zhou2023propainter} that rely on modern architectures like diffusion models. Moreover, Miao et al. \cite{miao2025rose} proposed an approach to tackle the removal of the objects as well as their associated effects. However, their method requires the collection of a large amount of synthetic data using 3D engines followed by an expensive training with this data. The closest to our work is Omnimatte-zero \cite{samuel2025omnimattezero} that attempts to remove the associated effects by identifying the corresponding tokens from the attention maps. Omnimatte-zero suffers from two major drawbacks. Firstly, Omnimatte-Zero constructs associated-effect masks by expanding from the user-provided object mask: it identifies regions that are strongly attended by tokens inside the object mask and adds them to the mask. However, this augmented mask is suboptimal, as it can miss the regions of associated effects with weaker activations and relies solely on the object mask as a seed. Secondly, it utilizes a heavy weight external model, TAP-Net \cite{doersch2023tapir} to track the foreground points and find associated background points in all the frames and further leverages these associations to compute the attention of the foreground points. This makes the attention computation for the foreground locations vulnerable to the inaccurate point tracking, especially in the cases of fast motion like a car speeding away, textureless areas or translucent objects. 

To overcome these drawbacks, we propose a two step approach to obtain the effects for the associated mask by first, leveraging the text-to-visual cross attention scores and identifying the visual tokens that are highly attentive to the query text tokens depicting the object to be removed and the associated effects. Given this set of seed visual tokens for the associated effects, we utilize the visual self attention scores to further refine this set and obtain the final mask that depicts the object and the associated effect. We relinquish the usage of any external model that may have introduced the erroneous computation of the attention for the foreground. Instead, we reinitialize the foreground region with Gaussian noise and, during the early denoising steps, when the global structure is formed, we bias the attention in the foreground region towards the background tokens using attention scaling. In the later steps, which mainly refine details, we just let the denoising process proceed normally, yielding an appropriate filling in the mask region. We show that the holistic nature of traditional metrics like peak-signal-to-noise-ratio (PSNR) or video quality scores used to evaluate the object removal in videos have several limitations in that it is easy to score high even when the object is not at all removed or only partially removed. In order to address this issue, we propose a novel metric, \textbf{Tok}en \textbf{Sim}ilarity (\textbf{\ourmetric}) that is designed for the problem of object removal from videos. Specifically, our metric rewards the similarity between foreground tokens in consecutive frames, similarity between foreground and background tokens in the same frame and dissimilarity between the foreground tokens in the input video and the output video. To summarize, the contributions of our work are the following. 
\begin{itemize}

\item We propose a training-free approach, \textbf{\ourmethod} that removes objects and their associated effects by localizing the associated region by utilizing cross-attention and self-attention in MMDiT blocks (see Fig.~\ref{fig:teaser}). 

\item We introduce a timestep-adaptive masking strategy with foreground reinitialisation and attention scaling, which prevents object leakage during denoising and enables effective object removal.

\item Further, given the paucity of evaluation metrics for object removal, we devise a new object removal metric (\textbf{\ourmetric}) that rewards high-quality object removal and heavily penalizes partial-to-no object removal.
\item We introduce a new real-world benchmark with associated effects and evaluate on it and DAVIS, showing \textbf{\ourmethod} outperforms all baselines (including training-based) on \textbf{\ourmetric} while remaining competitive on traditional metrics.
\end{itemize} 
\section{Related Works}
\label{sec:related}
\noindent
\textbf{Video Inpainting:}
Image and Video inpainting gained prominence due to the success of patchmatch, graphcuts, and energy minimization-based algorithms \cite{barnes2009patchmatch, newson2014video, granados2012background, huang2016temporally} that operated at the pixel level. With the evolution of deep learning, a number of techniques \cite{kim2019deep, xu2019deep, zhang2019internal, chang2019free, hu2020proposal, zhou2023propainter} were developed that cast the video inpainting as a pixel-to-pixel transformation. However, unlike our proposed approach, all the above video inpainting techniques are entirely focused on removing the objects and not the associated effects. \\
\textbf{Training-free method for Image and Video Editing:}
With the rise of diffusion models \cite{rombach2022high, peebles2023scalable}, training-free editing models \cite{geyer2023tokenflow, ceylan2023pix2video, qi2023fatezero, cong2023flatten, kara2024rave} has gained prominence due to the inherent advantages of not having to finetune the pretrained models. They are primarily designed to make prompt-driven low-level edits (stylization, color changes) and are not suitable for high-level tasks like object removal. \\
\textbf{Object removal:} 
The emergence of powerful video generative models have enabled the development of the several object removal techniques. Techniques like Rose \cite{miao2025rose}, Diffueraser \cite{li2025diffueraser}, Videopainter \cite{bian2025videopainter} all rely on collecting large amounts of mask data for objects in every frame of the video and finetuning a diffusion based generative model for object removal. This suffers from the similar drawbacks that the associated effects are retained in the output videos while being data-intensive. Vace \cite{vace} proposed a unified framework for video editing tasks ranging from low-level colorization to high-level object removal and addition and is data and training intensive. Recently, training-free approaches have been proposed for removing objects from still images \cite{zhu2025kv}. As we show in experiments, adopting this approach as is for videos does not remove the associated effects and results in significant artifacts in background. The closest to our approach are zeropatcher \cite{yangzeropatcher} and omnimattezero \cite{samuel2025omnimattezero}. As mentioned earlier, Omnimatte-zero suffers from many drawbacks in that it uses an external model for point tracking to compute the associations of the foreground points in order to compute the foreground attention values while denoising. Different from this, we do not utilize any external model and compute the foreground attention through reinitialization and attention bias towards background tokens. Ominmatte-zero relies on the user-provided object mask to compute the associated effects that we found to be suboptimal. We, instead propose a novel approach to associated mask computation by leveraging the text-to-visual cross-attention and visual self-attention scores, and show the mask obtained is indeed superior to the one computed in Omnimatte-zero.

\section{Methodology}
Our goal is to remove not only the object but also its associated effects in a training-free paradigm. Our approach has three steps: 1) Associated Effects Localization, wherein we leverage the cross-attention and self-attention maps in conjunction with the query text tokens to localize the object and its associated effects, 2) Inversion of the input video latent to obtain structured noisy latent while saving some intermediate background values, computing timestep adaptive masks and performing attention scaling, 3) Denoising of the noisy latent with re-initialization of the object, copying back the background values and performing attention scaling. Given an RGB video sequence of $k$ frames, $\mathcal{I}_k$, a corresponding binary mask sequence $\textbf{M}^{obj}$ denoting the object to be removed in each frame and a pair of text prompts, \{$P_s$, $P_T$\} describing source and target video. The goal is to generate a video $\mathcal{\hat{I}}_k$ with both the object and its associated effects removed while preserving the background.

\begin{figure}[!t]
    \centerline{\includegraphics[width=\columnwidth]{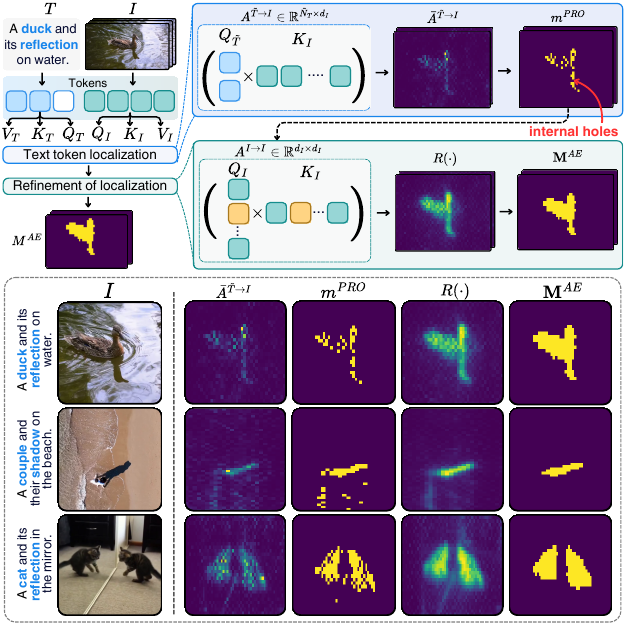}}
    \caption{Associated effects localization. The figure shows the processing of the video latents to obtain the mask for the object and the associated objects using the cross attention maps and the self attention maps. First, through the cross-attention scores, we obtain the patches of interest that are highly correlated with the query text tokens. Further, through self-attention scores, we identify the tokens that have the highest response to these patches of interest to obtain the final mask.}
    \label{fig:associated_effects}
\end{figure}

\begin{figure*}[t]
    \centering
    \includegraphics[width=\textwidth]{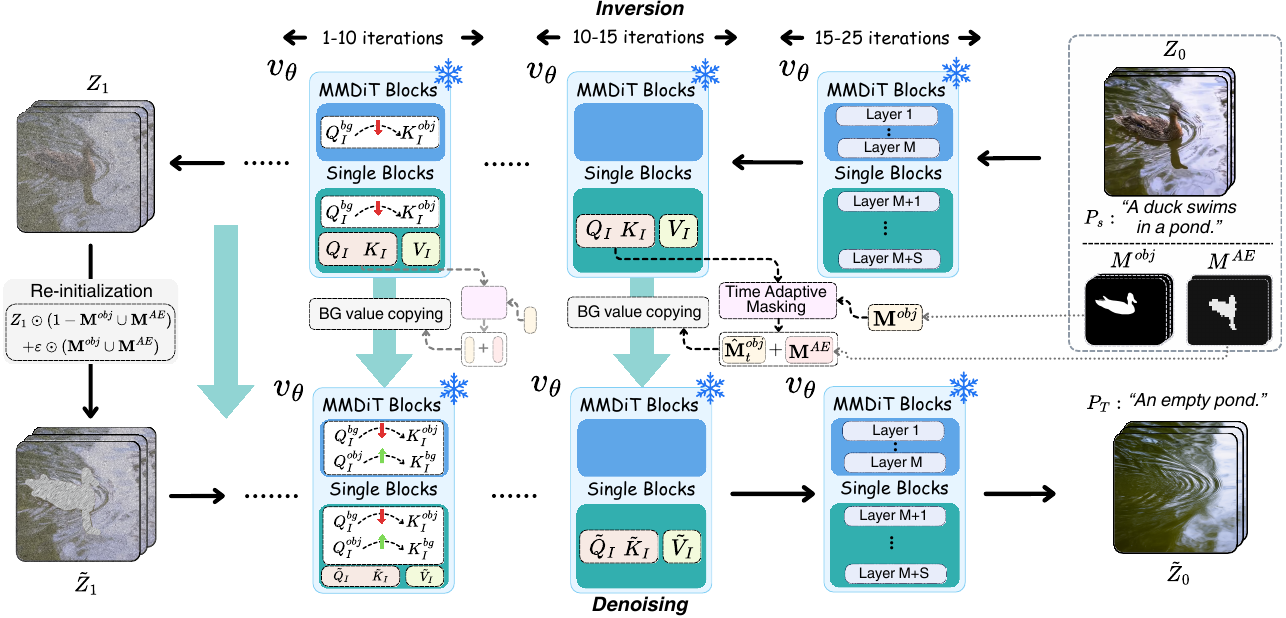}
    \caption{The figure shows all parts of our object removal algorithm once the mask for the associated effects algorithm is obtained. We perform inversion of video latent using RF Solver Edit while saving the background values for several iterations. The inverted noise is reinitialised in the mask region and is denoised with copying back the background values to obtain the output video.}
      \vspace{-5mm}
    \label{fig:obj_removal}
\end{figure*}
\subsection{Associated Effects Localization}
 
We aim to remove both the object and its associated effects (e.g., shadows, reflections, etc.). Since only the object mask is provided, we must augment it to cover both the object region and its associated-effect region. An overview of this module is shown in Fig.~\ref{fig:associated_effects}. Multi-modal DiT-based image and video generation models such as FLUX~\cite{flux2024}, Hunyuan~\cite{kong2024hunyuanvideo}, and CogVideoX~\cite{yang2024cogvideox} utilize joint attention (MM-DiT) layers that operate on a shared embedding space for text and visual tokens. We leverage this shared representation to localize visual tokens that correspond to both the object and its associated effects. 
The joint attention in MMDiT can be split into four components: text self-attention, visual self-attention ($I \to I$), cross-attention from text to visual tokens ($T \to I$) and cross-attention from visual to text tokens ($I \to T$). 

In any particular layer, the text features $\mathbf{f}_T \in \mathbb{R}^{N_T \times d_T}$ and the video features $\mathbf{f}_I \in \mathbb{R}^{N_I \times d_I}$ are projected into a shared embedding dimension $d$ as follows. 
\begin{align}
\mathbf{Q}_T &= \mathbf{f}_T\mathbf{W}_T^Q , \quad
\mathbf{K}_T = \mathbf{f}_T\mathbf{W}_T^K, \quad
\mathbf{V}_T = \mathbf{f}_T\mathbf{W}_T^V, \\
\mathbf{Q}_I &= \mathbf{f}_I\mathbf{W}_I^Q, \quad
\mathbf{K}_I = \mathbf{f}_I\mathbf{W}_I^K, \quad
\mathbf{V}_I = \mathbf{f}_I\mathbf{W}_I^V,
\end{align}
where $\mathbf{W}^{Q,K,V}_{T} \in \mathbb{R}^{d_T \times d}$ and $\mathbf{W}^{Q,K,V}_{I} \in \mathbb{R}^{d_I \times d}$ are projection matrices. 

\noindent \textbf{Query Text Token Based Localization:}
The goal is to identify the visual tokens that highly correlate to the object to be removed
(e.g., ``duck'') and its associated effect (e.g., ``reflection''). From the full set of text queries $\mathbf{Q}_T$, we extract $N_{\tilde{T}}$ subset of relevant tokens to get $\mathbf{Q}_{\tilde{T}} \in \mathbb{R}^{N_{\tilde{T}} \times d}$.
We leverage $T\rightarrow I$ cross attention to obtain attention map, $\mathbf{A}^{\tilde{T} \to I}$ using eq.~\ref{eq:cross_att}, indicating how strongly each visual token is linked to 
the object-related text queries.
\begin{equation}
    \mathbf{A}^{\tilde{T} \to I} = 
    \mathrm{Softmax}\!\left(
    \frac{ \mathbf{Q}_{\tilde{T}}\cdot \mathbf{K}_I^{\top} }
    { \sqrt{d} }
    \right)
\label{eq:cross_att}
\end{equation}
Averaging $\mathbf{A}^{\tilde{T} \to I}$ across the selected query tokens yields a single relevance map ($\bar{\mathbf{A}}^{\tilde{T} \to I} \in \mathbb{R}^{N_I} $) over visual tokens. We can reshape this text relevance map and visualize as shown in Fig.~\ref{fig:associated_effects} (top). 
Applying Otsu thresholding produces a proposal mask $m^{\text{PRO}}$. We observe that this mask is able to partially localize tokens of the object and its associated effects (see Fig.~\ref{fig:associated_effects} (top right)) but may still contain \textit{internal holes} when some relevant tokens receive weaker attention. We therefore treat $m^{\text{PRO}}$ as an initial proposal and refine it in the next stage 
using visual self-attention to obtain a dense, complete associated-effect mask.

\noindent \textbf{Self-Attention Based Refinement of Localization:}
Intuitively if the internal holes belong to the object of interest, then they must have high attention to the already identified tokens in the proposal mask $m^{\text{PRO}}$. To identify the `missing' tokens, we first obtain the visual self-attention map $\mathbf{A}^{I \to I} \in \mathbb{R}^{d_I \times d_I}$ given by eq. \ref{eq:self_att}.
\begin{equation}
    \mathbf{A}^{I \to I} = 
    \mathrm{Softmax}\!\left(
    \frac{ \mathbf{Q}_I\cdot \mathbf{K}_I^{\top} }
    { \sqrt{d} }
    \right) 
\label{eq:self_att}
\end{equation}
For each of the $N_I$ tokens, we compute the ratio of the sum of their attention values with respect to the proposal visual tokens and the sum of their attention values (in $\mathbf{A}^{I \to I}$) with respect to every visual token. This computation gives a response map, $\mathbf{R}(\cdot)$ that is further thresholded to obtain the final associated effect map, $\mathbf{M}^{AE}$ (see Fig.~\ref{fig:associated_effects}). 
In Fig.~\ref{fig:associated_effects} (bottom), we visualize $\mathbf{M}^{AE}$ and its intermediate stages for three videos. These examples demonstrate the robustness of our associated-effects localization module across diverse effect types.
$\textbf{M}^{AE}$ is used to union with the user-provided mask $\textbf{M}^{obj}$ and help remove associated effects. 

Unlike previous works~\cite{samuel2025omnimattezero,lee2025generative}, which use the input object mask $\textbf{M}^{obj}$ as $m^{\text{PRO}}$ gives us suboptimal results in comparison to our two-step approach. Additionally, only object text (i.e., only ``duck" token) or only associated effect text token (i.e., only ``reflection") is unable to provide our desired mask. Please refer to supplementary for more details.

\subsection{Inversion}
We adopt the inversion-denoising framework, which is widely used in training-free video-editing methods~\cite{jiao2025unieditflowunleashinginversionediting, wang2024taming}, for our training-free object removal approach. An overview of our approach is illustrated in Fig.~\ref{fig:obj_removal}. 

The source video latent $\mathbf{Z}_0$ is inverted to noise $\mathbf{Z}_1 \sim \mathcal{N}(0, I)$ using pre-trained text-to-video generation model, $v_\theta$, and source prompt $\textbf{P}_s$ using RF-Solver \cite{wang2024taming}. During the inversion, we store the attention features $\mathbf{V}_I$ in the last $r$ self-attention blocks and for last $k$ timesteps. 

\begin{figure}[t]
\centerline{\includegraphics[width=\linewidth]{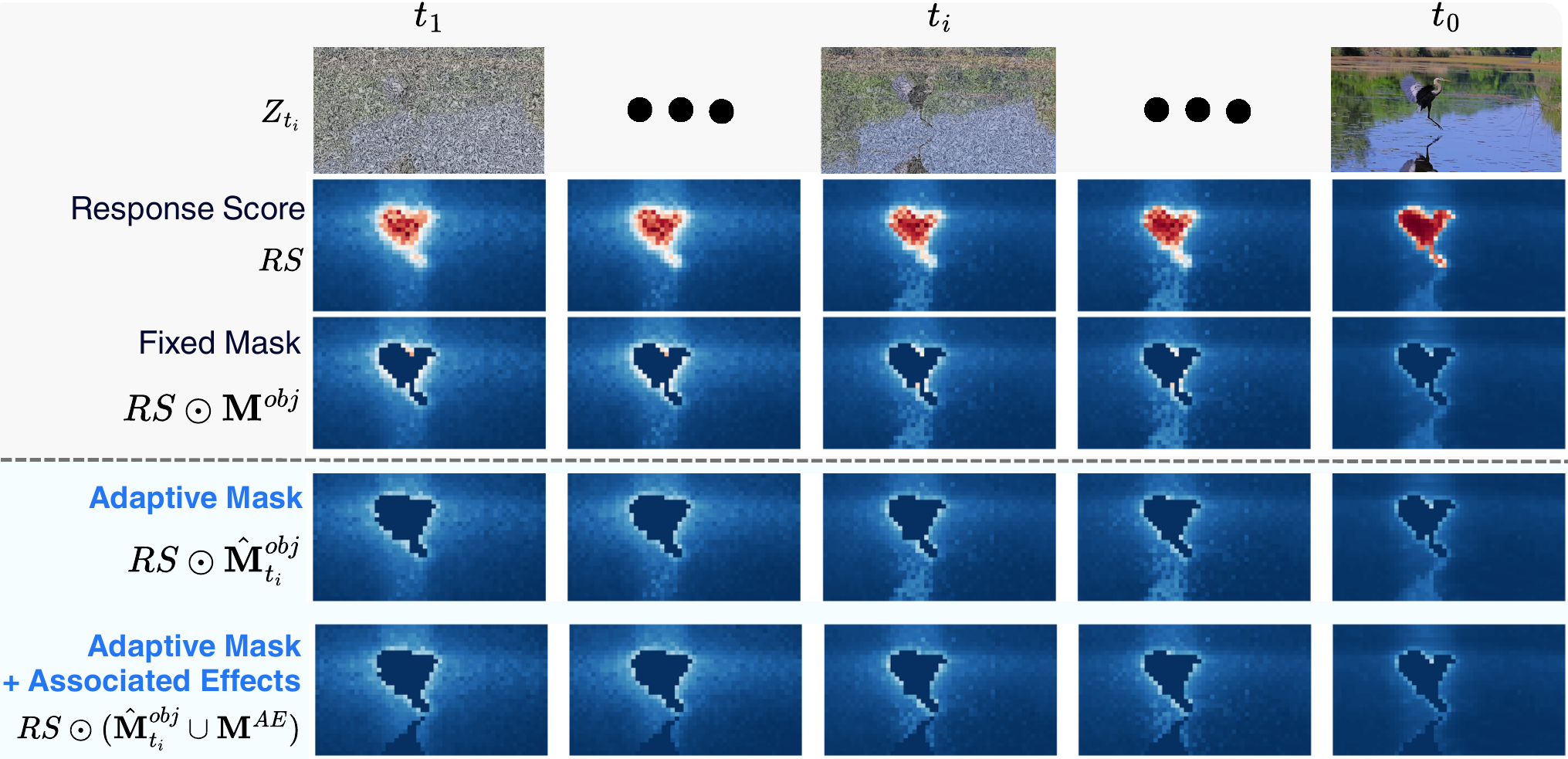}}
    \caption{Timestep adaptive masking. During inversion, the object’s footprint expands as noise increases, causing fixed masks to leak object tokens while denoising copying. In contrast, adaptive masks augmented with associated-effect regions prevent such leakage and enable complete removal of the object and its effects.}
    \label{fig:adaptive_mask}
\end{figure}

\noindent \textbf{Time Step Adaptive Masking:} To better understand the object presence in the attention space of video model, we analyze the self-attention layers in the model~\cite{lee2025generative}. We show this analysis in Fig.~\ref{fig:adaptive_mask}. For a fixed frame $j$ of the video latent $Z(j)$ we analyse the object presence at different timesteps $t_i$. Specifically, we measure the object response score ($RS$) of a query at the spatial location p (in the same frame) to the object at same frame ($\mathbf{M}^{obj}(j)$): 
\begin{equation}
    RS_p(j) = 
    \frac{
        \sum_{y \in \mathbf{M}^{obj}(j)} A^{I(j) \rightarrow I(j)}_{p,y}
    }{
        \sum_{x \in \mathcal{I}(j)} A^{I(j) \rightarrow I(j)}_{p,x}
    }.
\end{equation}
We observe that as we move closer to noisy distribution during inversion, the presence mask $RS(j)$ starts increasing. Due to self-attention through so many steps the object presence keeps on increasing. Most previous approaches use a fixed mask through time and if we overlay the object mask $\textbf{M}^{obj}$ on $RS(j)$ (see row 3 in Fig.~\ref{fig:adaptive_mask}), we observe that fixed mask actually is not able to fully cover the object region. Hence we update the mask by thresholding the $RS(j)$ to get $\hat{M}_t^{obj}$ map. We calculate this during inversion for the last $t_i \in [k-1,0]$ timesteps using all the self-attention layers (as shown in Fig.~\ref{fig:obj_removal} for $k$=15 timesteps during inversion ). Similar to value features, we store these Adaptive mask indices. In the presence of associated effect, we also add the $\textbf{M}^{AE}$ mask to the adaptive mask. We see in row 4 and 5 in Fig.~\ref{fig:adaptive_mask}, how adaptive masking and adding $\textbf{M}^{AE}$ covers object and associated mask. This will be used to skip copying values during the denoising step corresponding to the object and its associated effect to better remove the object. 

\noindent \textbf{Attention Scaling:} Along with recognising the object relevant video tokens, while inverting we also want the background to integrate less information from the object (and it associated effect). Specifically, using the mask ($\mathbf{M}^{obj} \cup \mathbf{M}^{AE}$) we can divide $\mathbf{Q}_I$,$\mathbf{K}_I$,$\mathbf{V}_I$ tokens to get object relevant $\mathbf{Q}^{obj}_I$,$\mathbf{K}^{obj}_I$,$\mathbf{V}^{obj}_I$ and background relevant $\mathbf{Q}^{bg}_I$,$\mathbf{K}^{bg}_I$,$\mathbf{V}^{bg}_I$. 
\begin{equation}
    \tilde{\mathbf{A}}^{bg \to obj} = 
    \mathrm{Softmax}\!\left(
    \frac{\mathbf{Q}^{bg}_I\cdot (c \mathbf{K}^{obj}_I)^{\top} }
    { \sqrt{d} }
    \right),
    \label{eq:attn_scale}
\end{equation}
where $c < 1$. We only apply this to last few timesteps of inversion and to all the layers (as shown in Fig.~\ref{fig:obj_removal} for last 10 timesteps).
Note that since we estimate the time-adaptive mask using the $\mathbf{Q}_I$,$\mathbf{K}_I$ of current timestep, we do not have access to $\hat{\mathbf{M}}_t^{obj}$ during inversion. 

\subsection{Denoising}
\noindent \textbf{Reinitialization:} The inversion process maps the source video $\mathbf{Z}_0$ to a noise latent $\mathbf{Z}_1$ in gaussian distribution. $\mathbf{Z}_1$ contains the structural and semantic information corresponding to $\mathbf{Z}_0$. Similar to KV-Edit~\cite{zhu2025kv}, we reinitialise the object region with gaussian noise ($\varepsilon$). 
$\tilde{\mathbf{Z}}_1 = \mathbf{Z}_1 \odot (1-\mathbf{M}^{obj} \cup \mathbf{M}^{AE}) + \varepsilon \odot (\mathbf{M}^{obj} \cup \mathbf{M}^{AE})$
Note that here the masks are at the latent shapes. Essentially, reinitialization removes any prior about the object and its associated effect from the latent and want the model to inpaint this region with the background information. During denoising, we start from a noisy $\tilde{\mathbf{Z}}_1$ latent, containing noisy background prior and no object prior, and prompt $\mathbf{P}_T$, our aim is to reconstruct the background as closely as possible to source video and infill or construct the object region with plausible information. 

\noindent \textbf{Attention Scaling:} Since the object region is randomly initialised and do not have any semantic or structural information, we explicitly rely on the background tokens to fill appropriate object region. Specifically we modify the $\mathbf{A}^{obj \to bg}$ attention with 
\begin{equation}
    \tilde{\mathbf{A}}^{obj \to bg} = 
    \mathrm{Softmax}\!\left(
    \frac{\mathbf{Q}^{obj}_I\cdot (b \mathbf{K}^{bg}_I)^{\top} }
    { \sqrt{d} }
    \right),
\end{equation} where $b > 1$. 
Given our goal is to reconstruct the background, similar to inversion we update $\mathbf{A}^{bg \to obj}$ using eq.~\ref{eq:attn_scale}. Unlike inversion, during denoising we have access to more accurate $\hat{\mathbf{M}}_t^{obj} \cup \mathbf{M}^{AE}$ mask to separate object and background relevant tokens. Since the structure is formed during the initial timesteps, we applying the attention scaling during the first few timesteps and on all layers (as shown in Fig.~\ref{fig:obj_removal} for first 10 timesteps). Similar to other training-free editing methods~\cite{wang2024taming,zhu2025kv}, we also copy the background value features. For the last $r$ layers of single block, suppose $\tilde{\mathbf{V}}_I$ is the value feature. We copy value features from $1-\hat{\mathbf{M}}_t^{obj} \cup \mathbf{M}^{AE}$.
In later timesteps, we let the model denoise naturally, blending the inpainted and background regions into a coherent video.
\section{\ourmetric: Object Removal Metric}

\begin{figure}[t]
    \centering
    \includegraphics[width=0.95\linewidth]{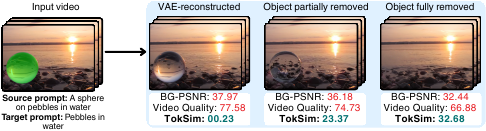}
    \caption{The proposed metric, \ourmetric~scores very high only when the object is fully removed and progressively becomes lower as the object removal reduces. For VAE-reconstruction where the object is not removed at all, the \ourmetric~is nearly zero. However, the ranges of the values for BG-PSNR and video quality across the vastly different outputs are extremely compressed and do not serve the purpose of unambiguously distinguishing between the object removal approaches of varied capabilities.}
    \label{fig:token_metric}
\end{figure}

Previous object removal approaches~\cite{liu2025generative, zhu2025kv} compare using metrics like background-PSNR (bg-PSNR) and quality.
The inherent limitation of these metrics is that they are not designed for object removal and can easily circumvent the actual removal task while still obtaining high metric values.
For example in Fig.~\ref{fig:token_metric}, if there exists an algorithm that produces the output video that is nearly the same as the input video (for eg. encode-decode using a VAE), bg-PSNR and Quality are high while the object is very apparently present. 
While text-alignment can be used for semantic alignment, it does not account for temporal consistency in videos. 

Motivated by these shortcomings of the existing metrics for object removal,
we propose \textbf{Tok}en \textbf{Sim}ilarity metric, \textbf{\ourmetric}, a metric to evaluate object removal in videos. Given the rich semantic and structural information provided by the DINOv3~\cite{simeoni2025dinov3}, we operate at token-level. Given input video, object mask and predicted video (from an object removal method), {\ourmetric} (i) rewards temporal consistency in the object tokens in consecutive frames, (ii) penalizes the similarity between the object patches of the input video and the output video and (iii) rewards similarity between the object and the neighbouring background tokens. Intuitively, a good object removed video should sit well with the background and time and should be far from the original object. Specifically, using DINOv3~\cite{simeoni2025dinov3} and object mask we extract frame-wise object and background token embeddings for both input and output video. For each object token at location $z$ in frame $k$, we compute its cosine similarity with the token at the same location in frame $k+1$ to obtain $\lambda_{z}^{k}$. Similarly, for each object token at location $z$ in frame $k$ in the output, we compute its cosine similarity with the corresponding token in the input to obtain $\eta_{z}^{k}$.
In addition, for each foreground token at location $z$ in frame $k$, we compute the mean of its cosine similarities with nearby background tokens in the same frame to obtain $\tau_{z}^{k}$.
The final \ourmetric score for a given video is the mean of all object tokens and frames, given as in eq. \ref{eq:tokensim}.
\begin{equation}
 \text{\ourmetric} =  100 \cdot \frac{1}{F}\sum_{z=0}^{F-1}\sum_{i=1}^{N^{obj}} \lambda_{z}^{k} \cdot (1-\eta_{z}^{k}) \cdot \tau_{z}^{k}
\label{eq:tokensim}
\end{equation}
We can observe in Fig.~\ref{fig:token_metric} that unlike other metrics, {\ourmetric} can distinguish between object not removed, partially removed, and completely removed.
For videos with associated effects, we use the associated mask $\textbf{M}^{AE}$ along with the pixel mask. The key feature of the proposed {\ourmetric} metric is that if the object region is temporally coherent (high $\lambda$), is fully removed (high $1-\eta_{z}$), blends well with the background (high $\tau_{z}$) in the output video, it will score high. And similarly, if either the object region is not coherent or is not fully removed or does not blend well with the background, we will end up with a smaller {\ourmetric} value.
\begin{table}[t]\centering
    \renewcommand{\arraystretch}{0.5}
    \setlength{\tabcolsep}{4pt}
    \resizebox{\linewidth}{!}{
    \begin{tabular}{l|ccccccc|c}
        \toprule
        \bf Name & \bf Real & \bf Ref. & \bf Sh. & \bf Mir. & \bf T. & \bf M.E. & \bf D.A. & \bf \# \\
        \midrule
        DAVIS~\cite{perazzi2016benchmark} & \cmark & \cmark & \cmark & \xmark & \xmark & \xmark & \xmark & 50 \\
        Movies~\cite{lin2023omnimatterf} & \xmark & \cmark & \cmark & \xmark & \xmark & \xmark & \xmark & 5 \\
        Kubric~\cite{wu2022d} & \xmark & \xmark & \cmark & \xmark & \xmark & \xmark & \cmark & 5 \\
        GenProp~\cite{liu2025generative} & \cmark & \cmark & \cmark & \xmark & \xmark & \xmark & \xmark & 15 \\
        ROSE-Bench~\cite{miao2025rose} & \xmark & \cmark & \cmark & \cmark & \cmark & \xmark & \cmark & 60 \\
        \midrule
        \rowcolor{Light}
        \bf \ourbenchmark~(Ours) & \cmark & \cmark & \cmark & \cmark & \cmark & \cmark & \cmark & 60 \\
        \bottomrule
    \end{tabular}}
    \caption{Comparison with previous object removal benchmarks. Where Real=all real data, Ref.=Reflections present, Sh.=shadows present, Mir.=mirror, T.=translucent object, M.E.=multiple-simultaneous associations, D.A.=disconnected association, \#=number of videos.}
    \label{tab:new_bench}
\end{table}

\begin{figure*}[ht!]
    \centering
    \includegraphics[width=\textwidth]{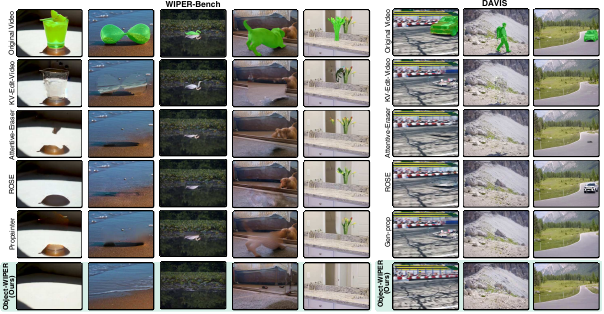}
    \caption{Qualitative comparison between our method and existing approaches on (left) WIPER-Bench and (right) DAVIS. On WIPER-Bench, our method removes both the object and its associated effects across diverse scenarios, whereas both training-free and training-based baselines fail to remove the object completely. On DAVIS, our method achieves full object removal; notably, in the car example (third column), even training-based methods such as Gen-Prop and ROSE are unable to do so.}
    \label{fig:qualitative_misc}
\end{figure*}

\begin{table*}[t]\centering
    \renewcommand{\arraystretch}{1.1}
    \setlength{\tabcolsep}{2pt}
    \resizebox{\textwidth}{!}{
    \begin{tabular}{ll|ccccc|ccccc}
        \toprule
         & \multirow{2.5}{*}{\bf Model} & \multicolumn{5}{c}{\bf DAVIS} & \multicolumn{5}{|c}{\bf \ourbenchmark} \\
         \cmidrule(lr){3-7}
         \cmidrule(lr){8-12}
         & & \bf \ourmetric$\uparrow$ & \bf BG-PSNR$\uparrow$ & \bf FG-flicker$\downarrow$ & \bf Text-align($\times 10^2$)$\uparrow$ & \bf Qual.($\times 10^2$)$\uparrow$ & \bf \ourmetric$\uparrow$ & \bf BG-PSNR$\uparrow$ & \bf FG-Flicker$\downarrow$ & \bf Text-align($\times 10^2$)$\uparrow$ & \bf Qual.($\times 10^2$)$\uparrow$ \\
        \midrule
        & VAE (Image)~\cite{flux2024} & 0.32 & 34.05 & 31.68 & 23.26 & 73.88 & 0.42 & 36.39 & 6.77 & 24.69 & 72.58 \\
        & VAE (Video)~\cite{kong2024hunyuanvideo} & 1.25 & 30.27 & 31.15 & 22.84 & 72.04 & 0.86 & 35.27 & 6.37 & 24.71 & 70.27 \\
        \midrule
        \multirow{4.5}{*}{\rotatebox{90}{ \demph{w/ Training}}} & {\demph{Propainter~\cite{zhou2023propainter}}}  & \demph{28.24} & \demph{34.01} & \demph{13.73} & \demph{26.18} & \demph{64.53} & \demph{20.99} & \demph{41.07} & \demph{4.00} & \demph{25.70} & \demph{68.88} \\
        & {\demph{ROSE~\cite{miao2025rose}}} & {\demph{29.36}} & {\demph{26.97}} & {\demph{18.89}} & {\demph{26.13}} & {\demph{61.59}} & {\demph{30.02}} & {\demph{30.90}} & {\demph{2.94}} & {\demph{26.70}} & {\demph{61.94}} \\
        & {\demph{VACE~\cite{vace}}} & \demph{15.86} & \demph{24.48} & \demph{22.83} & \demph{24.01} & \demph{63.76} & \demph{11.53} & \demph{29.67} & \demph{6.03} & \demph{25.24} & \demph{65.53} \\
        & {\demph{Gen-Prop~\cite{liu2025generative}}}\footnotemark{} & \demph{30.52} & \demph{24.27} & \demph{13.21} & \demph{25.89} & \demph{51.43} & \demph{-} & \demph{-} & \demph{-} & \demph{-} & \demph{-} \\
        \midrule
        \multirow{4.5}{*}{\rotatebox{90}{w/o Training}}
        & KV-Edit~\cite{zhu2025kv} & 23.17 & \bf 32.31 & 21.31 & 25.21 & \bf 65.83 & 14.46 & \bf 35.17 & 9.20 & 25.32 & \bf 67.25 \\
        & Attentive-Eraser~\cite{sun2025attentive} & 30.82 & 28.07 & 18.46 & 26.31 & 46.68 & 25.28 & 32.01 & 8.92 & 26.07 & 56.59 \\
        & KV-Edit-Video~\cite{zhu2025kv} & 28.68 & 25.78 & 18.32 & 25.21 & 59.91 & 23.26 & 31.76 & 4.74 & 25.70 & 63.69 \\
        \cmidrule{2-12}
        & \CC{Light}\textbf{\ourmethod~(Ours)} & \CC{Light}\bf 32.80 & \CC{Light}23.02 & \CC{Light}\bf 16.37 & \CC{Light}\bf 26.63 & \CC{Light}61.62 & \CC{Light}\bf 33.09 & \CC{Light}27.53 & \CC{Light}\bf 3.02 & \CC{Light}\bf 26.91 & \CC{Light}61.80 \\
        \bottomrule
    \end{tabular}
    }
    \caption{Quantitative comparisons. We compare Object-WIPER with prior training-based and training-free object removal methods. Object-WIPER achieves superior performance on the TokSim metric across both benchmarks, surpassing even training-based approaches.}
    \vspace{-5mm}
    \label{tab:main_table}
\end{table*}

\section{Experiments}
\noindent \textbf{Datasets:} 
The typical datasets that are utilized to benchmark object removal methods are shown in Tab.~\ref {tab:new_bench}. Existing video object removal datasets (i.e., DAVIS \cite{perazzi2016benchmark} and Genprop \cite{liu2025generative}) are limited to shadows and reflection types of associated effects. To evaluate the associated effects, previous works have relied on simulation-based data \cite{miao2025rose}. We introduce a new dataset, \textbf{\ourbenchmark}, made of only real videos curated from Pexels \cite{pexels} and Youtube \cite{youtube} and cover a wide set of associations - shadows, reflections (from reflective surfaces like water), mirrors and translucent objects, complex associations like simultaneous multiple associations and spatially disconnected object and effect associations. {\ourbenchmark} consists of $60$ videos that are $2$-second long and are collected at 24 frames per second (FPS) and each video is of resolution, either $480 \times 848$ or $720 \times 400$. Additionally, we use SAM2 \cite{ravi2024sam} to generate the masks of the objects to be removed (examples from {\ourbenchmark} in suppl.). Besides our dataset, we also conduct experiments on the DAVIS dataset \cite{perazzi2016benchmark} that consists of videos that have difficult scenarios like fast motion. 

\begin{figure*}[t!]
    \centering
    \vspace{-1em}
    \includegraphics[width=0.965\textwidth]{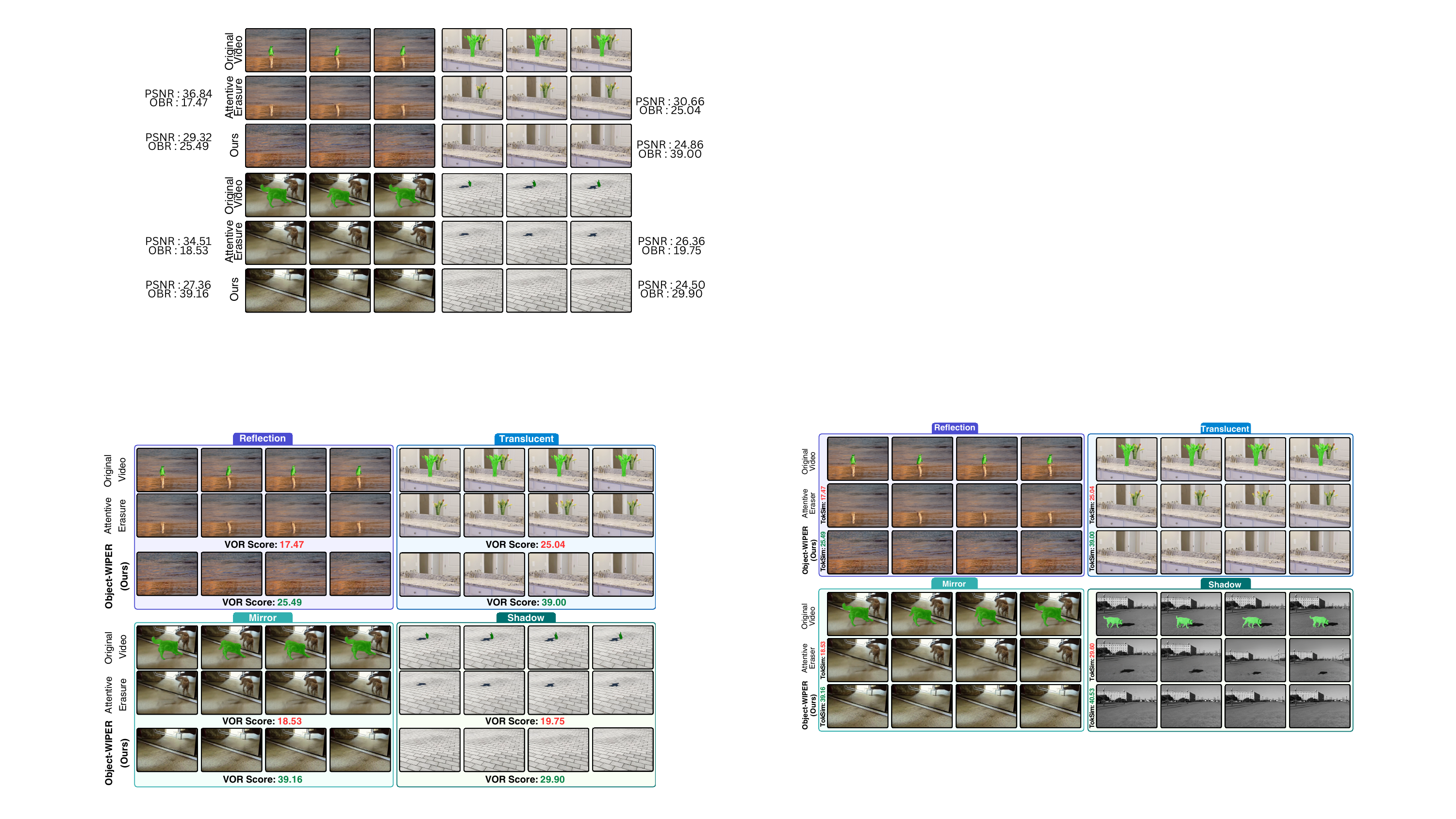}
    \caption{The figure shows the qualitative comparison of the proposed method and the best training-free baseline, Attentive Erasure \cite{sun2025attentive} for four different associated effects along with the \ourmetric Scores in each of the cases. In all these cases, unlike Attentive Erasure, our method clearly removes the object as well as the associated effects.}
    \vspace{-5mm}
    \label{fig:qualitative}
\end{figure*}

\noindent \textbf{Baselines and Metrics:} We compare our method with a state-of-the-art video inpainting method, Propainter \cite{zhou2023propainter}, training based diffusion model approaches like Genprop \cite{liu2025generative} and ROSE \cite{miao2025rose}, frame-wise training-free methods like KV-Edit \cite{zhu2025kv} and attentive eraser \cite{sun2025attentive} and training-free methods designed for single images but adapted for video, KV-Edit-Video \cite{zhu2025kv}. We were unable to compare with OmniMatte-Zero~\cite{samuel2025omnimattezero} due to the unavailability of public code. We resize the videos to fit into the dimensions expected by the models. We additionally compare with reconstructed video using FLUX frame-wise VAE and Hunyuan VAE for reference. Along with our new object removal metric, \textbf{\ourmetric}, we also evaluate the different methods using several metrics typically used such as background peak signal-to-noise ratio (BG-PSNR), foreground temporal flickering score (FG-Flicker) \cite{huang2024vbench}, text-alignment score (Text-align) \cite{radford2021learning} and DOVER score \cite{wu2023exploring} for Video Quality Score (Qual.). More details on baselines, metrics and implementation details in the Supplementary.

\begin{figure}[tb]
    \centering
    \includegraphics[width=0.47\textwidth]{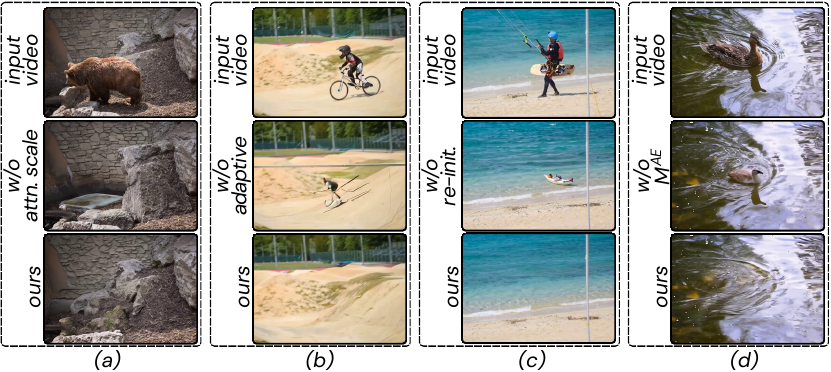}
    \caption{Qualitative ablation results. We remove each component of our model to assess its contribution: (a) attention scaling improves the coherence of the filled region, (b) timestep-adaptive masking enables removal of fast-moving objects, (c) reinitialization eliminates residual structures, and (d) the $M^{AE}$ mask removes the object together with its associated effects.}
    \label{fig:ablation_qual}
\end{figure}

\noindent \textbf{Qualitative results:}
In Fig.~\ref {fig:qualitative_misc}, we show several examples from the two datasets and compare the object removal results across all the baselines. \ourmethod is consistently able to remove the masked object as well as its associated effects. In particular, we highlight that even in the presence of translucence and shadow (first two examples in WIPER-Bench), only our method is able to remove the object, the shadow as well as fill the region with appropriate background. None of the existing methods, training-free or otherwise are able to able handle the mirrored objects. Genprop, which is the best training-based method, while removing the associated effects in some cases (first and last examples in DAVIS), fails in some cases of fast object motion, leaving remnants of the object in the output video. In Fig.~\ref{fig:qualitative}, we compare the output frames from our method against the best training-free baseline, Attentive Eraser for different associated effects, reflection, translucent, mirror and shadow. We also show the TokenSim scores for both methods. We can see that while our method is able to remove both the object as well as the associated effects whereas Attentive Eraser fails completely in removing the associated effect. 

\noindent \textbf{Quantitative Results:}
Tab.~\ref{tab:main_table} shows object removal performance of different methods as evaluated by different metrics on the two datasets. {\ourmethod} despite being training-free outperforms all baselines in terms of TokSim metric including the training based approaches like ROSE, Gen-Prop that are fine-tuned for associated effects removal. VAE reconstructions yield significantly low TokenSim scores as the object is not removed and the output DINOv3 features are nearly the same as the input DINOv3 features. However, for all other metrics, the difference between VAE reconstructions and the best method for that metric are much closer thus highlighting a clear deficiency of these metrics. We also note that the proposed method is based to score quite high on FG-flickering metric thus indicating high temporal consistency post object removal. Similarly, our method has high text-alignment scores indicating a high per-frame object and associated effects removal rate. 

\footnotetext{We sought results from the authors and could obtain only for DAVIS.}
\section{Discussion}

In Tab.~\ref{tab:ablation}, we show quantitatively how each component of our pipeline contributes to the object removal and qualitatively in Fig.~\ref{fig:ablation_qual}. Attention biasing forces the background to attend less to foreground during inversion and foreground to attend more to the background during denoising. This results in the removed region to be more homogenous with respect to the background and gives a boost of 1.1 dB in BG-PSNR. In cases where large objects are removed, the attention scaling is particularly effective as it reduces the dependence on the foreground values. We observe that adaptive masking helps in the cases when the object to be removed has high motion. Adaptive masking helps to avoid copying the image patch values that the usual mask would have missed and hence properly removing the object (see Fig.~\ref{fig:ablation_qual} (b)). Only with use of associated mask, $\mathbf{M}^{AE}$, the method is able to remove the associated effect (see Fig.~\ref{fig:ablation_qual} (d)). We also perform an ablation experiment on the various options that we have for masks and show why the chosen the union of time adaptive object mask and the computed associated effect mask works the best (see suppl. for more details).

\begin{table}[t!]\centering
    \renewcommand{\arraystretch}{0.9}
    \setlength{\tabcolsep}{4pt}
    \resizebox{\linewidth}{!}{
    \begin{tabular}{l|ccc}
        \toprule
          & \bf \ourmetric$\uparrow$ & \bf BG-PSNR$\uparrow$ & \bf Text-align($\times 10^2$)$\uparrow$ \\
        \midrule
        \rowcolor{Light}
        \textbf{\ourmethod} & \underline{32.80} & 23.02 & \textbf{26.63} \\
        \midrule
        w/o attention scaling & \textbf{32.97} & 21.92 & 26.42 \\
        w/o adaptive mask & 32.10 & 22.73 & \underline{26.44} \\
        w/o re-initialization & 30.36 & \textbf{23.47} & 25.92 \\
        w/o $\mathbf{M}^{AE}$ & 32.18 & \underline{23.10} & 26.17 \\
        \bottomrule
    \end{tabular}
    }
    \caption{Ablation of model components on DAVIS. Using all the components we get a good balance of object removal and background preservation and text alignment. }
    \label{tab:ablation}
\end{table}

\section{Conclusion}
We propose \textbf{\ourmethod}, a training-free method for removing objects and their associated effects from videos using a state-of-the-art text-to-video diffusion model. Our method can localize the associated effects based on the cross-attention and self-attention scores in the DiT blocks, given the query text depicting the object and the effect. Through examples, we show that the existing metrics are not suitable to evaluate the ability of an object removal algorithm and hence to overcome this, we propose a novel metric, {\ourmetric} that rewards approaches that remove the objects cleanly and penalizes approaches that remove only partially. Through quantitative and qualitative experiments, we show that our training-free approach beats all methods including training-based approaches in terms of the proposed metric. 
\clearpage

\noindent \textbf{Acknowledgments.} We thank Tianyu Wang and Soo Ye Kim for helpful discussions on GenProp. We also thank Anugya Shah for assistance with annotating WIPER-Bench, and Preet Mukeshkumar Sojitra for help with setting up and running some baseline experiments.

{
    \small
    \bibliographystyle{ieeenat_fullname}
    \bibliography{main}
}

\clearpage
\setcounter{page}{1}
\maketitlesupplementary

\addtocontents{toc}{\protect\setcounter{tocdepth}{2}} 
\setcounter{tocdepth}{2}

\hypersetup{linkcolor=black}
\tableofcontents

\section{WIPER-Bench}
\subsection{Dataset construction details}
We collected videos from Pexels~\cite{pexels} and YouTube~\cite{youtube} by searching for keywords such as ``\textit{shadow}", ``\textit{reflection}", ``\textit{mirror}", ``\textit{translucent}", ``\textit{transparent}", ``\textit{animal + shadow/reflection}" and ``\textit{object + shadow/reflection}". We avoided videos where a person's face was clearly visible, to maintain privacy and ethical reasons. In addition to simple scenes, we also included complex videos containing disconnected associated effects or multiple co-occurring effects. In total, we manually downloaded 52 candidate videos. From each video, we selected at most two non-overlapping 2-second clips, resulting in 74 candidate samples.

All landscape videos were resized to a resolution of $480 \times 848$, and portrait videos were resized to $720 \times 400$. We also resampled all videos to 24\,fps. For annotation, we manually labeled the object masks frame-by-frame using the SAM2~\cite{ravi2024sam} demo interface. A few videos resulted in huge segmentation errors when SAM2 was applied and were therefore discarded. After balancing category distribution, our final dataset consists of 60 videos.

\subsection{Examples and statistics}
\begin{figure}[h!]
    \centerline{\includegraphics[width=0.48\textwidth]{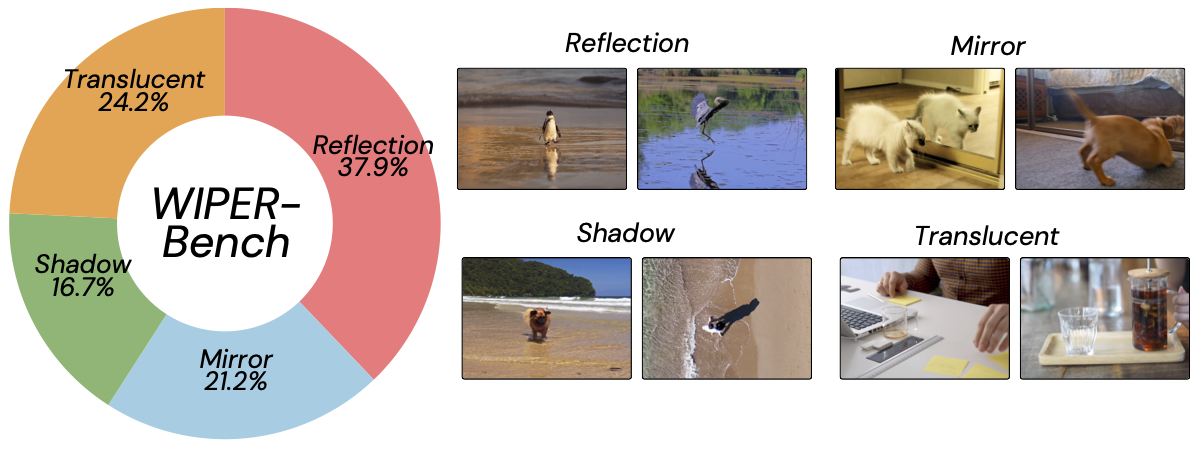}}
    \caption{Statistics and example cases from WIPER-bench for evaluating object removal with associated effects.}
    \label{fig:our_bench}
\end{figure}

\begin{figure}[h!]
    \centerline{\includegraphics[width=0.48\textwidth]{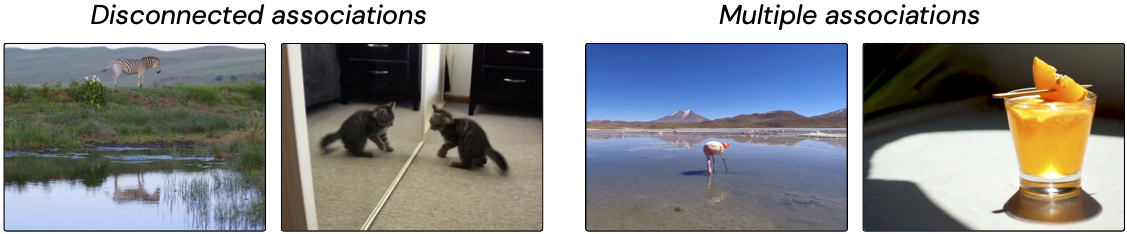}}
    \caption{WIPER-bench also includes naturally occurring complex cases, such as disconnected associations and multiple co-occurring associations.}
    \label{fig:our_bench_more}
\end{figure}

Given the collected data, the distribution of categories is shown in Fig.~\ref{fig:our_bench}. These statistics reflect the natural availability of such phenomena in real-world videos. The final dataset includes 25 reflection cases, 14 mirror cases, 11 shadow cases, and 16 translucent associated effects. Additionally, 6 videos contain multiple associated effects, and 12 videos include disconnected associations. Examples of multiple and disconnected associations are shown in Fig.~\ref{fig:our_bench_more}.

\section{Implementation details}

\subsection{Object-WIPER model details}
We use pretrained Hunyuan-T2V model~\cite{kong2024hunyuanvideo} as our video-generation model. It consists of $M=20$ MMDiT and $S=40$ single blocks. We the use RF-Solver~\cite{wang2024taming} sampler for inversion and denoising that has $25$ time steps through the model. We store and copy background feature values for $k=15$ time steps and last $r=20$ single (or self-attention ). We use classifier free guidance (cfg) value of $1$ during inversion and $5$ during denoising. We apply adaptive masking for $k=15$ time steps and using all $40$ single blocks. For all the MMDiT and single blocks, we apply attention scaling for $10$ steps. We choose $c=0.8$ and $b=1.2$. To calculate the associated mask we use MMDiT layers of intermediate time steps $t_i \in \{6,7,10\}$. To improve readability, we summarize all symbols and notations used in the paper in Tab.~\ref{tab:notation}.

\begin{table*}[t]
\centering
\caption{Summary of notations used throughout the paper.}
\label{tab:notation}
\resizebox{0.99\textwidth}{!}{
\begin{tabular}{llll}
\toprule
\textbf{Variable} & \textbf{Value} & \textbf{Dimension} & \textbf{Description} \\
\midrule
$\mathcal{I}_k$ & - &  $ 3 \times (F+1) \times H \times W $  & Input pixel video frames \\
$\hat{\mathcal{I}}_k$  & - &  $ 3 \times (F+1) \times H \times W $  & Predicted pixel video frames \\
$\mathbf{Z}_{t}$ & - & $ 16 \times (F/4+1) \times H/8 \times W/8 $ & Video latent at timestep $t$ during inversion \\
$\tilde{\mathbf{Z}}_{t}$ & - & $ 16 \times (F/4+1) \times H/8 \times W/8 $ & Video latent at timestep $t$ during denoising \\
$\mathbf{Z}(j)$ & - & $ 16 \times 1 \times H/8 \times W/8 $ & $j^{th}$ video latent frame during inversion \\
\midrule
\multirow{3}{*}{$\mathbf{M}^{obj}$} & - & $ 1 \times (F+1) \times H \times W $ & User provided binary object pixel mask  \\
 & - & $ 16 \times (F/4+1) \times H/8 \times W/8 $ & Max-pool \& repeat to align with video latent \\
 & - & $ 1 \times (F/4+1) \times H/16 \times W/16 $ & Max-pool to align with video tokens \\
\midrule
\multirow{3}{*}{$\mathbf{M}^{AE}$} & - & $ 1 \times (F/4+1) \times H/16 \times W/16 $ & Estimated associated mask aligned with video tokens \\
 & - & $ 16 \times (F/4+1) \times H/8 \times W/8 $ & Upsampled \& repeat to align with video latent \\
 & - & $ 1 \times (F+1) \times H \times W $ & Upsampled binary mask to align with pixel video \\
 \midrule
$\hat{\textbf{M}}^{obj}_t$ & - & $ 1 \times (F/4+1) \times H/16 \times W/16 $ & Estimated adaptive mask at timestep $t$ aligned with video tokens \\
$m^{PRO}$ & - & $ 1 \times (F/4+1) \times H/16 \times W/16 $ & Estimated Proposal mask at timestep aligned with video tokens \\
$P_s$; $P_t$ & - & - & Input source and target text prompts, respectively \\
$\mathbf{f}_T$; $\mathbf{f}_I$ & - & - & Video and text feature embeddings, before attention \\
$N_I$; $N_T$ & - & - & Number of video patches and text tokens \\
$d_T$; $d_I$; $d$ & - & - & Video feature dimension; Text feature dimension; Shared dimension \\
$(\mathbf{Q}_T,\mathbf{K}_T,\mathbf{V}_T) ; (\mathbf{Q}_I,\mathbf{K}_I,\mathbf{V}_I)$ & - & $N_T \times d$; $N_I \times d$ & Query, Key \& Values for Video and Text tokens \\
$\mathbf{A}^{X \to Y}$ & - & $N_X \times N_Y$ & Attention maps from X to Y ($X,Y \in \{I,T\}$ ) \\
$RS(j)$ & - & $ 1 \times (F/4+1) \times H/16 \times W/16 $ & Object response score for $j^{th}$ frame aligned with video token \\
$c$ & 0.8 & - & Attention scaling factor for background to object attention  \\
$b$ & 1.2 & - & Attention scaling factor for object to background attention  \\
$k$ & 15 & - & Number of timesteps for value feature saving (inversion) and copying (denoising) \\
$r$ & 20 & - & Number of last single blocks for which value saving and copying happens \\
\bottomrule
\end{tabular}
}
\end{table*}

\subsection{Baseline details}

\noindent \textbf{Training based methods:} We compare our method against several state-of-the-art object removal approaches, including VACE \cite{vace}, ProPainter \cite{zhou2023propainter}, ROSE \cite{miao2025rose}, and GenProp \cite{liu2025generative}. ROSE and GenProp are trained to remove both object and its associated effect, similar to it we want to do that in a training-free way. For VACE, ProPainter, and ROSE, we use the official checkpoints and publicly released implementations. As GenProp is not open-source, we contacted the authors directly and obtained their predicted videos for evaluation.

\noindent \textbf{Training-free methods.} Given our training-free approach, we mainly compare our method with previous (open-sourced) training-free approaches, including KV-Edit~\cite{zhu2025kv} and Attentive-Eraser. Since these approaches are image-based we implement for the video by running them frame-wise. We extend KV-Edit for videos, as explained next.

\noindent \textbf{KV-Edit-Video}
KV-Edit~\cite{zhu2025kv} demonstrates strong performance on image-based object removal and is originally implemented on the FLUX~\cite{flux2024}. However, performing object removal independently on each frame does not account for temporal consistency in videos. Given the architectural similarity between FLUX and Hunyuan, and to ensure a fair comparison, we extend KV-Edit to operate on the Hunyuan video model.

Following their approach, we store all intermediate tokens and (self-attention) video key/value features during inversion. We then reinitialize the tokens corresponding to the object region and, during denoising, replace the tokens and (self-attention) key/value features for the background region with those saved from inversion. Due to CPU memory limitations, we exclude saving and restoring the key/value tensors for the MMDiT blocks.

\begin{figure}[h!]
    \centerline{\includegraphics[width=0.48\textwidth]{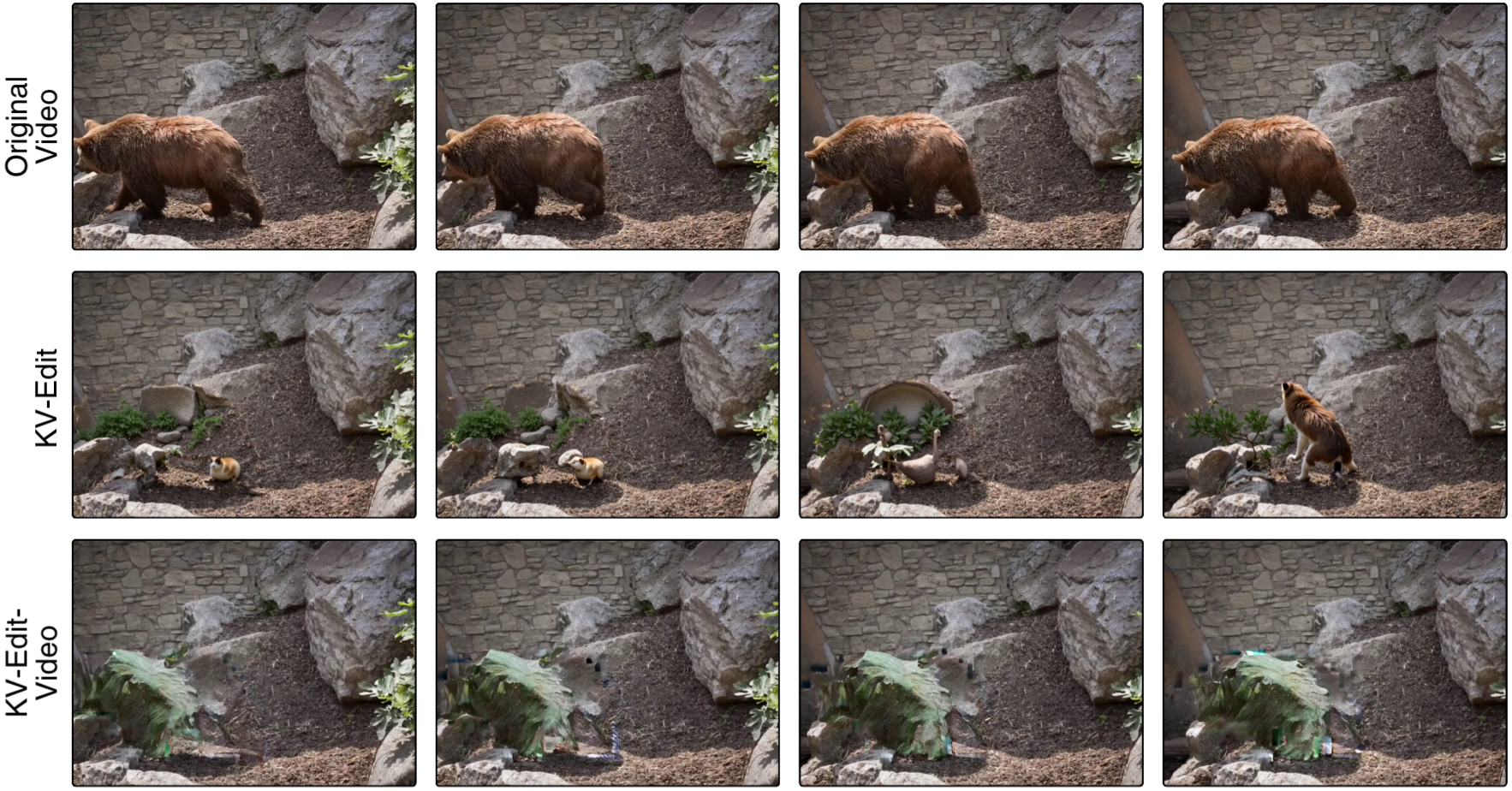}}
    \caption{Object removal comparison. KV-Edit (frame-wise) produces temporally inconsistent inpainting across frames. Extending the method to video latents, KV-Edit-Video, improves temporal coherence, but this extension still introduces noticeable artifacts along object–background boundaries.}
    \label{fig:kv_edit}
\end{figure}

We illustrate an example of object removal in Fig.~\ref{fig:kv_edit}. The (frame-wise) KV-Edit produces inpainted regions that are temporally inconsistent across frames. Extending KV-Edit to operate on video tokens improves temporal coherence in the inpainted regions. 
However, KV-Edit-Video still introduces boundary inconsistencies and noticeable artifacts because it copies background tokens and attention features using a fixed mask. In contrast, our method employs a timestep-adaptive masking strategy that refines the fixed mask avoids copying all background tokens, resulting in both temporally and spatially consistent object removal.

\noindent \textbf{OmnimatteZero} 
OmnimatteZero~\cite{samuel2025omnimattezero} introduces a training-free approach for generating video omnimattes. One of their intermediate goals involves removing foreground objects to get backgrounds. However, due to the unavailability of public code and insufficient implementation details, we were unable to reproduce their method and therefore could not include it in our comparisons. Moreover, their primary focus is on producing omnimattes and evaluating them on simulated datasets specifically designed for that task.

In contrast, our objective is to remove objects from real-world videos and to evaluate performance directly on such real data. Unlike omnimatte datasets, which provide ground-truth background videos without objects, real videos do not have ground-truth object-free references. To address this gap, we also propose a new evaluation metric, \textbf{TokSim}, tailored for assessing object removal quality in real-world videos.


\subsection{Evaluation metric details.}
\noindent \textbf{TokSim.} 
Due to the lack of appropriate metrics for evaluating object removal in videos, we propose TokenSimilarity, a token-level metric computed using image patch embeddings extracted from DINOv3. For each pair of consecutive frames $f$ and $f{+}1$, we first compute the union of their object masks. 
If the object has been successfully removed, the union of the masks defines the object-token region, which should now resemble the surrounding background tokens and remain consistent with the corresponding region in the next frame.

For the tokens within the object region, we measure their embedding distance to the corresponding tokens in the ground-truth frame $f$, as well as their similarity to tokens in frame $f{+}1$. Additionally, we compare these object-region tokens with nearby background patches $f_{\text{bg}}$) within a 24-pixel neighbourhood outside the union mask. These comparisons collectively quantify how well the removed region integrates with its temporal and spatial context.

\noindent \textbf{BG-PSNR.}
We evaluate background preservation by computing the PSNR (Peak Signal-to-Noise Ratio) over the unmasked regions of the video.

\noindent \textbf{FG-flickering.}
Temporal flickering was introduced in VBench~\cite{huang2024vbench} to assess the temporal quality of generated videos. Building on this idea, we compute the L1 difference between consecutive frames, but restrict the evaluation to the object region. For each pair of consecutive frames, we take the union of their object masks and compute the L1 distance only within this region. By focusing on the former object area, FG-flickering isolates the temporal stability of the inpainted region, making it significantly more sensitive to object-removal inconsistencies than global flicker metrics.

\noindent \textbf{Text-alignment.} 
We compute the cosine similarity between the CLIP~\cite{radford2021learning} embeddings of the output video frame and the target text prompt.

\noindent \textbf{Quality.} 
We use DOVER~\cite{wu2023dover} to measure overall video quality. However, we observe that this global metric does not reliably reflect the quality of object removal.

For videos containing associated effects, we expand the original object mask by taking its union with the upsampled (calculated) associated-effect masks. This augmented mask more accurately separates the object ( {+} associated effect ) region from the background for evaluation.

\section{Limitations}
While our method is particularly impressive in identifying the associated effects and removing them, we note that the inherent nature of the training-free paradigm in which our method operates in introduces several limitations. Specifically, background preservation ability of our method is limited by the reconstruction ability of the RF-Solver Edit~\cite{wang2024taming}. For example, the background PSNR of the inversion–denoising reconstruction on the DAVIS dataset is only 25.44\,dB. This indicates that even RF-Solver Edit alone can introduce undesirable artifacts in the background region during inversion and denoising.

Our approach is further bounded by the capacity of the underlying video diffusion model and its VAE reconstruction. The video model may struggle with highly complex or previously unseen cases, leading to degraded results. Notably, the background PSNR of the Video-VAE reconstruction on DAVIS (30.27\,dB) is 3.7\,dB lower than that of the Image-VAE reconstruction (34.05\,dB), highlighting a gap in reconstruction quality that directly impacts background preservation of our approach.


\section{User studies}
\subsection{Interface and setup} 
\begin{figure}[h!]
    \centerline{\includegraphics[width=0.48\textwidth]{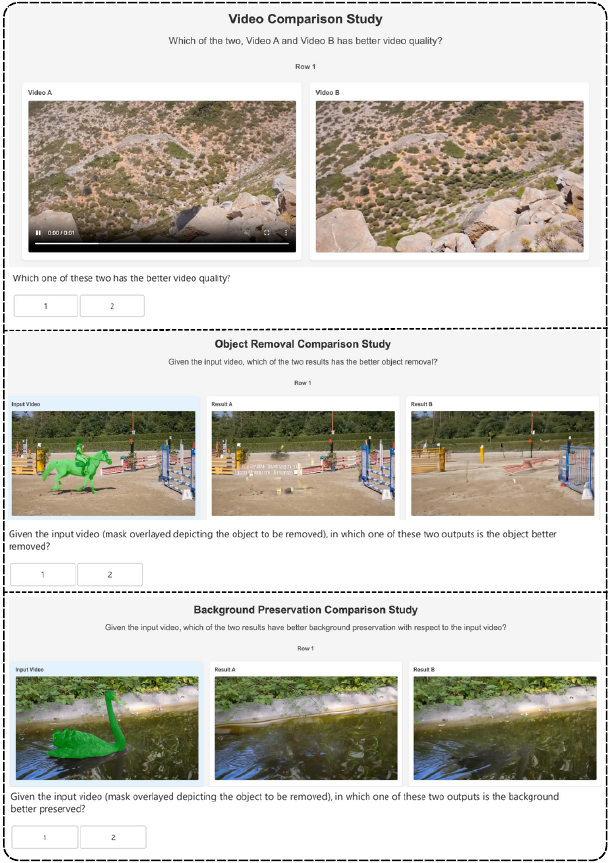}}
    \caption{User study interface. We ask the users three types of questions related to video quality, object removal quality and background preservation quality.}
    \label{fig:user_study}
\end{figure}

We conduct human evaluation study to show the efficacy of our method in the training-free regime as well as the effectiveness of \ourmetric in estimating the object removal ability of different methods. Specifically, we do $15$ pairwise comparisons between our result and a baseline result randomly selected from one of the three training-free algorithms, KV-edit \cite{zhu2025kv}, KV-edit-video \cite{zhu2025kv} and Attentive Eraser \cite{sun2025attentive}, for three separate questions, `Video Quality', `Object Removal' and `Background Preservation'. We show the interface for user-study in Fig.~\ref{fig:user_study}.
For the video quality assessment, we show only the results from our approach and one of the baseline approaches and ask the question, `Which of the two videos has better video quality?'. For the object removal assessment, we show the input video with the mask for the object to be removed overlayed and the two results, and ask the question, `Given the input video, which of the two results have better object removal?'. For the background preservation study, we show the input video with mask for the object to removed overlayed and the results, and ask the question, `Given the input video, which of the two results have better background preservation with respect to input video?'.
\subsection{Analysis}
\paragraph{Human Preferences:}
In total we collected responses from 10 users across 45 pairwise comparisons, making it a total of 450 responses. For video quality, our method was preferred 96.67\% of the times. For the object removal, our method was preferred 90.67\% of the videos, and for background preservation, our method was preferred 77.33\% of the times. As shown through metrics in the main paper, it is expected that our method performs betters in terms of video quality, object removal as opposed to background preservation.
\paragraph{TokSim and Human Preference Agreement:} We also obtained TokSim for each of the videos in the pairwise comparisons and determined which video was preferred if we strictly assume higher TokSim scores is akin to better object removal. We dub these as `TokSim Preferences'. For each of the 15 pairwise comparisons, we compare the TokSim preferences with perferences of 10 users and found that TokSim preferences is 83.64\% accurate with respect to human. This clearly shows the value of using the metric proposed in being a strong replacement of human evaluation.
\paragraph{Inter-rater Agreement:} Inter-rater reliability was assessed using Fleiss $\kappa$ which is appropriate for evaluating consistency among more than two raters who assign categorical judgments \cite{fleiss1973equivalence}. The observed $\kappa$ value of 0.72 indicates substantial agreement amongst the raters suggesting that they demonstrated a high level of concordance in their evaluations and that the ratings are sufficiently consistent to support subsequent analyses \cite{landis1977measurement}.

\section{Associated Effects Localization details}
Since only the object mask is provided and the associated effects also need to be removed, we leverage the model's prior knowledge encoded in the unified text--video token space within the joint-attention (MMDiT) layers. For reflection and shadow cases, we use text tokens corresponding to both the \textit{object} and its \textit{effect} to guide the removal process. For mirror cases, where the reflected object is visually real object, we found that using only the \textit{object}-related text tokens yields better localization.

\subsection{Analysis on text tokens}
\begin{figure}[h!]
    \centerline{\includegraphics[width=0.48\textwidth]{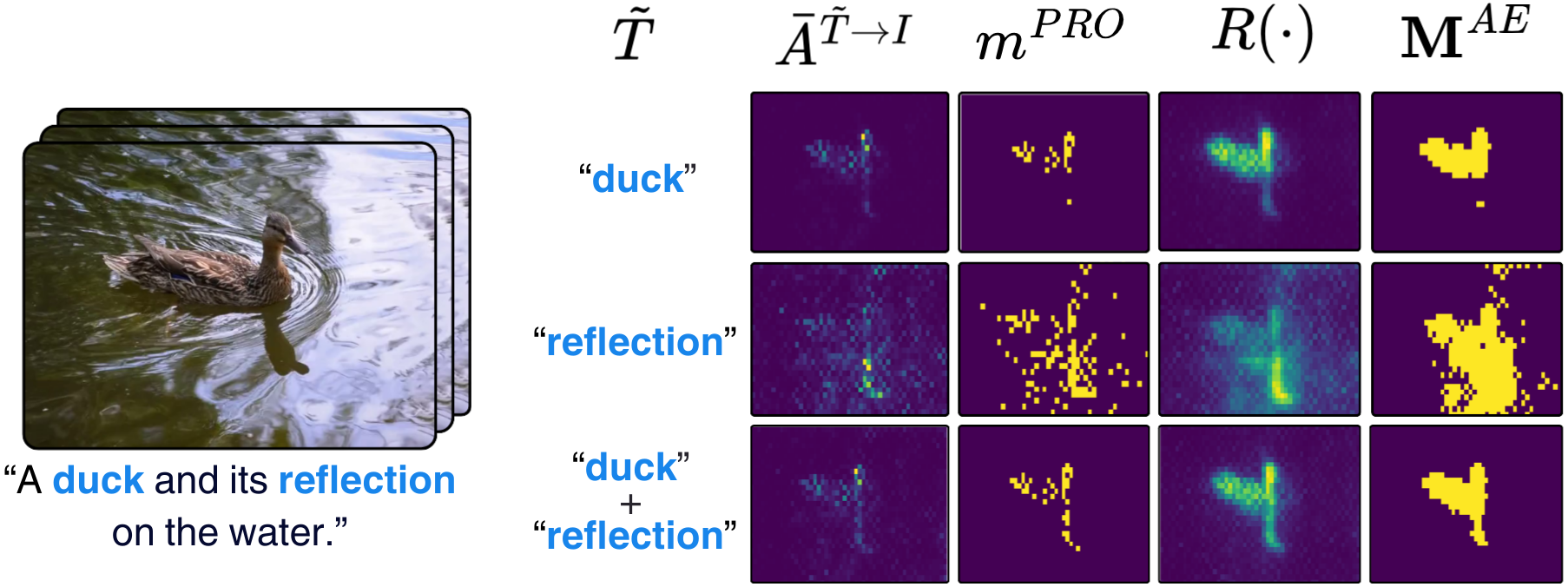}}
    \caption{Effect of text tokens on localization. Using only \textit{object}-text tokens or only \textit{effect}-text tokens leads to incorrect localization, whereas combining both yields accurate object–effect masks.}
    \label{fig:ae_text}
\end{figure}

For shadow and reflection associated effects, we empirically find that using only \textit{object}-text tokens or only \textit{effect}-text tokens fails to capture the full object–effect region. For example, as shown in Fig.~\ref{fig:ae_text}, using only ``duck'' text tokens highlights only the object, while using only ``reflection'' tokens produces incorrect and overly spread localizations. Therefore, we jointly use both token types, which yields a compact and accurate localization of the object and its associated effect.


\subsection{Replacing $m^{PRO}$ with $\textbf{M}^{obj}$}
We analyse whether the proposal mask $m^{PRO}$ must be computed using text guidance, or if the user-provided object mask alone can serve as an adequate proposal. As shown in Fig.~\ref{fig:ae_ablate}, skipping the proposal-mask estimation step results in masks that fail to capture the associated effects. This highlights the importance of the text-guided proposal stage for associated effect localization.

\subsection{Limitation of $\textbf{M}^{AE}$ using OmnimatteZero}
\begin{figure}[h]
    \centerline{\includegraphics[width=0.45\textwidth]{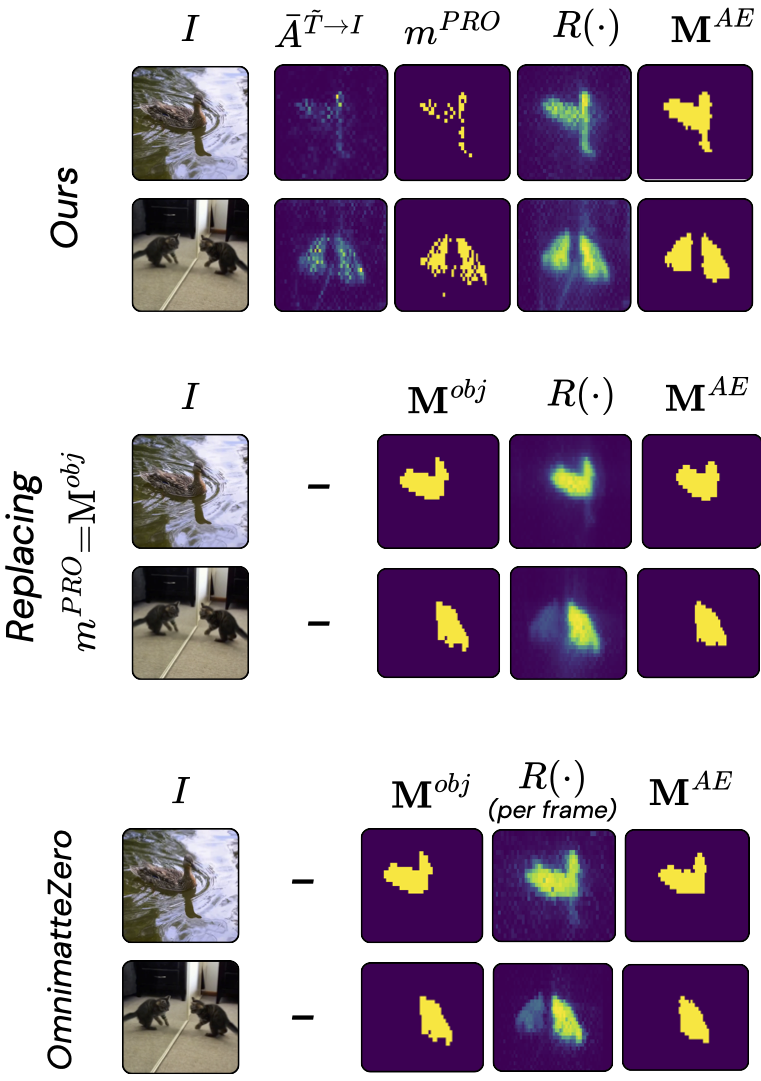}}
    \caption{Comparison of associated-effect mask localization. \textbf{Top:} Our method accurately localizes both the object and its associated effects. \textbf{Middle:} Replacing $m^{PRO}$ with the user-provided $\mathbf{M}^{obj}$ (i.e., skipping the proposal-mask estimation) results in masks that fail to capture the associated effects. \textbf{Bottom:} Approaches used by OmnimatteZero~\cite{samuel2025omnimattezero} and Generative-Omnimatte~\cite{lee2025generative} are unable to correctly localize the associated-effect regions.}    
    \label{fig:ae_ablate}
\end{figure}

Generative-Omnimatte~\cite{lee2025generative} and OmnimatteZero~\cite{samuel2025omnimattezero} estimates the associated-effect regions by selecting per-frame high-response tokens conditioned on the user-provided object mask $\mathbf{M}^{obj}$. However, as shown in Fig.~\ref{fig:ae_ablate}, this strategy fails to correctly identify the associated-effect regions.

\subsection{Limitation of Concept attention}

\begin{figure}[h!]
    \centerline{\includegraphics[width=0.48\textwidth]{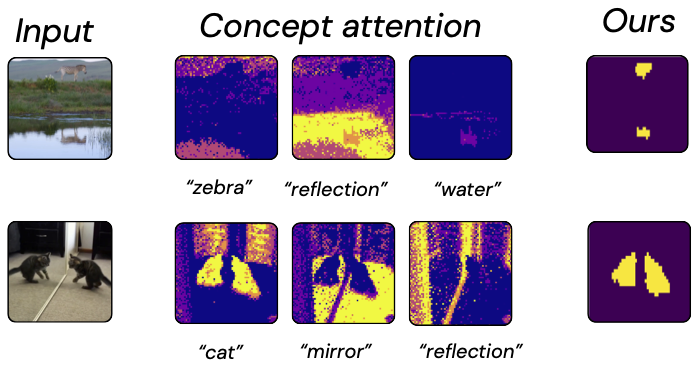}}
    \caption{Comparison with concept attention~\cite{helbling2025conceptattentiondiffusiontransformerslearn}. We show the (left) input image and (middle) activations for text concepts using Concept attention and (right) our estimated object-associated effect. We observe that concept-attention struggle to precisely localize the object and its associated effects, while our text-to-video approach provides accurate localization.}
    \label{fig:concept}
\end{figure}

We observe that text‐to‐image approaches~\cite{helbling2025conceptattentiondiffusiontransformerslearn, hu2025dcedit}, which use text prompts to localize concepts in images, struggle to achieve the level of spatial precision required to distinguish the object, its associated effects, and the background. As shown in Fig.~\ref{fig:concept}, concept attention often produces coarse or ambiguous activations that fail to correctly isolate both the object and its associated effects. This makes it non-trivial to leverage such methods for accurate object (+associated effect) separation from background.
In contrast, our text‐to‐video–based approach provides significantly sharper and more consistent localization, enabling reliable identification of both the object and its associated effects.

\subsection{Ablation on masks.} 
We compare how would the combination of different masking strategy helps. In Tab.~\ref{tab:ablation_mask}, we compare on the subset of DAVIS with associated effects. We observe that our strategy of Adpative masking on $M^{obj}$ and adding $M^{AE}$ outperforms any other combination of masking for object removal. 

\begin{table}[h]\centering
    \renewcommand{\arraystretch}{0.9}
    \setlength{\tabcolsep}{4pt}
    \resizebox{\linewidth}{!}{
    \begin{tabular}{l|ccc}
        \toprule
         \bf Masking Strategy & \bf \ourmetric$\uparrow$ & \bf BG-PSNR$\uparrow$ & \bf Text-align($\times 10^2$)$\uparrow$ \\
        \midrule
         $\mathbf{M}^{obj}$ & 27.69 & \underline{22.11} & 25.69 \\
         $\mathbf{M}^{AE}$ & 26.75 & 21.69 & 25.16 \\
         $\mathbf{M}^{obj} \cup \mathbf{M}^{AE}$ & \underline{28.66} & 21.63 & 25.84 \\
         $\hat{\mathbf{M}}_t^{obj}$ & 27.19 & \textbf{22.37} & 25.56 \\
         $\hat{\mathbf{M}}_t^{AE}$ & 28.52 & 21.99 & \underline{26.20} \\
         $\widehat{(\mathbf{M}^{obj} \cup \mathbf{M}^{AE})}_t$ & 28.49 & 21.64 & 26.00 \\
         \midrule
         \rowcolor{Light}
        $\hat{\mathbf{M}}_t^{obj} \cup \mathbf{M}^{AE}$ \textbf{(Ours)} & \textbf{29.32} & 21.64 & \textbf{26.54} \\
        \bottomrule
    \end{tabular}
    }
    \caption{Ablation on DAVIS subset with associated effects. $\mathbf{M}^{obj}$, $\mathbf{M}^{AE}$, $\hat{\mathbf{M}}_t(\cdot)$ are the object, associated and time adapted mask, respectively.}
    \label{tab:ablation_mask}
\end{table}

\section{Running time comparison}
We compare the runtime of our method against training-free baselines. For fairness, we exclude model-loading and I/O overheads (image/video loading and saving) and report only the inference time. The results are averaged over 10 runs on videos of size $25 \times 480 \times 848$ (Frames $\times$ Height $\times$ Width) and shown in Tab.~\ref{tab:run_time}. As shown in Tab.~\ref{tab:run_time}, our method achieves inference time comparable to existing training-free approaches, while surpassing them in object-removal quality.

\begin{table}[h!]\centering
    \renewcommand{\arraystretch}{0.9}
    \setlength{\tabcolsep}{4pt}
    \resizebox{0.6\linewidth}{!}{
    \begin{tabular}{l|c}
        \toprule
         \bf Method & \bf Run-time$\downarrow$ (sec)  \\
        \midrule
         KV-Edit\cite{zhu2025kv} & 323.85 \\
         Attentive-Eraser & 305.52 \\
         KV-edit-Video & 551.35 \\
         \midrule
         Object-WIPER (ours) & 354.69 \\
        \bottomrule
    \end{tabular}
    }
    \caption{Run-time comparison. Our method achieves comparable inference time}
    \label{tab:run_time}
\end{table}

\end{document}